\documentclass{article} 

\title{A Theory of Feature Learning}

\author{
Brendan van Rooyen\\
Department of Computer Science\\
The Australian National Unversity/NICTA\\
\texttt{Brendan.VanRooyen@nicta.com.au} \\
\And
Robert C. Williamson\\
Department of Computer Science\\
The Australian National Unversity/NICTA\\
\texttt{Bob.Williamson@nicta.com.au} \\
}

\usepackage{nips14submit_e,times}
\usepackage{url}
\usepackage{bbold,amsmath,bm,amsthm,amssymb}
\usepackage{float}
\usepackage{graphicx}
\usepackage{grffile}
\usepackage[all]{xy}
\usepackage{booktabs}
\usepackage{times}

\newcommand{\RR}{\protect\mathbb{R}}
 
\newcommand{\EE}{\protect\mathbb{E}}

\newcommand{\dist}{\raise.17ex\hbox{$\scriptstyle\mathtt{\sim}$}}
\newcommand{\Lbar}{\underline{L}}
\newcommand{\Rbar}{\underline{R}}

\DeclareMathOperator*{\arginf}{arg\,inf}

\theoremstyle{plain}
\newtheorem{theorem}{Theorem}
\newtheorem*{theorem*}{Theorem}

\newtheorem{lemma}[theorem]{Lemma}

\theoremstyle{definition}
\newtheorem{definition}[theorem]{Definition}

\nipsfinalcopy

\begin{document}
\maketitle
\thispagestyle{empty}

\begin{abstract}
Feature Learning aims to extract relevant information contained in data sets in an automated fashion. It is driving force behind the current deep learning trend, a set of methods that have had widespread empirical success. What is lacking is a theoretical understanding of different feature learning schemes. This work provides a theoretical framework for feature learning and then characterizes when features can be learnt in an unsupervised fashion. We also provide means to judge the quality of features via rate-distortion theory and its generalizations.
\end{abstract}

\section{Introduction}

Machine Learning methods are only as good as the features they learn from. This simple observation has led to a plethora of feature learning methods. From methods that aim to learn features and a linear classifier in one go such as neural networks and predictive sparse coding \cite{Mairal2012,Bradley2008,Hinton2006}, to methods based on conditional independence tests \cite{Tishby2000,Wang2010,Fukumizu2011,Banerjee2005}, to unsupervised feature learning methods \cite{Hinton2006,Olshausen1997,Guyon2009,Berkhin2006,Vincent2008} and of course good old fashion hand engineered features. While there exist many heuristic justifications for these methods, what is lacking is a general theory of feature learning. 
$$
\xymatrix{
\mbox{data} \ar[r] & \boxed{\mbox{Feature\ Map}}  \ar[r] & \boxed{\mbox{Classifier}} \ar[r] & \mbox{prediction}
}
$$
We are all familiar with the above flow chart. Many methods exist to each of the above components. For a real application we are interested in measuring the predictive performance of the combined system. For the sake of \emph{understanding} we seek means to measure the quality of each \emph{component}. Thus we seek a measure of the quality of a feature map that is \emph{independent} from the rest of the overall system, as well as a means to \emph{combine} this with the generalization performance of a classification algorithm to provide bounds on the overall performance of the entire system.

To this end we review both supervised and unsupervised feature learning schemes, presenting a novel supervised feature learning algorithm as well as novel transfer of regret bounds results. We draw inspiration from both rate distortion theory \cite{Cover2012} as well as the comparison of statistical experiments \cite{Cam2011,Torgersen1991}. We provide to our knowledge the \emph{first} framework from which to understand feature learning as well as a \emph{characterization} (theorem 5) of when unsupervised feature learning is possible within our framework. Our characterization \emph{metrizes} feature learning, in the sense that we give means to calculate the amount of information lost by any feature map. We show how many existing schemes for feature learning can be understood as surrogates to theorem 5. Finally we show how rate-distortion theory can be used to rank the quality of features.

\section{Notation and Preliminaries}

Throughout the paper $Y,X,Z$ and $A$ will denote the label, instance, feature and action spaces respectively. We allow $A$ to be arbitrary to included both classification and conditional probability estimation amongst others. $L$ will denote a loss function $L: Y \times A \rightarrow \RR_+$. Denote by $\lVert L \rVert = \sup_{y,a} \lvert L(y,a) \rvert$ the norm of the loss. Furthermore for two sets $X$ and $Y$ the set of all functions $f: X \rightarrow Y$ will be denoted by $Y^X$.

For a set $X$ denote the set of probability distributions on $X$ by $\mathcal{P}(X)$. Denote by $\lVert P -Q \rVert$ the variational divergence between $P$ and $Q$ \cite{Reid2009b}, a standard metric on probability distributions. 

Define a \emph{Markov kernel} \cite{Morse1966} from a set $X$ to a set $Y$ to be a measurable function $P_{Y|X} : X \rightarrow \mathcal{P}(Y)$, in the sense that for all measurable $f: Y \rightarrow \RR$ we have $f^*(x) = \EE_{y \dist P_{Y|X}(x)} f(y)$ is a measurable function. Markov kernels provide means to work with conditional probability distributions. As shorthand $P_{Y|X}(x) = P_{Y|x}$. All measurable functions $f: X \rightarrow Y$ define Markov kernels, with $P_{Y|x} = \delta_{f(x)}$. Denote by $M(X,Y)$ the set of all Markov kernels from $X$ to $Y$.

Given two Markov kernels $P_{X|Y}$ and $P_{Z|X}$ we can \emph{compose} them to form $P_{Z|X} \circ P_{X|Y}: Y \rightarrow \mathcal{P}(Z)$, essentially by marginalizing out $X$ in the Markov chain $Y \rightarrow X \rightarrow Z$ \cite{Torgersen1991,Morse1966}. One has 
$$
\EE_{P_{Z|X} \circ P_{X|y}} f =  \EE_{ x \dist P_{X|y}} \EE_{z \dist P_{Z|x}} f(z)
$$ 
for all measurable $f: Z \rightarrow \RR$. Given a Markov kernel $P_{Y|X}$ and a distribution $P_X \in \mathcal{P}(X)$ we can form a joint distribution $P_{XY} = P_{X} \otimes P_{Y|X}$ in the standard way. Similarly by Bayes rule we have $P_{XY} = P_{X} \otimes P_{Y|X} = P_{Y} \otimes P_{X|Y}$. Such a ``disintegration" holds for very general measure spaces \cite{Chang1997,Simmons2012}.

We assume that learning follows the protocol: First, nature draws $(x,y) \dist P_{XY}$. Second, the learner observes $x$ and chooses an action $a$. Finally, the learner incurs loss $L(y,a)$. We view the loss function as an integral part of the learning problem. We place no restrictions on its form. We refer to $P_{X|Y}$, the class conditional distributions, as the \emph{experiment}.

Let $P \in \mathcal{P}(Y)$ and $L$ be a loss. Define the \emph{Bayes act} $a_P := \arginf_a \EE_{y \dist P}L(y, a)$. If multiple Bayes acts exist then pick one of them. For many cases of interest there is always a \emph{unique} Bayes act. This is true for all strictly proper losses \cite{Reid2009b,Parry2012} as well as kernel mean based losses \cite{Grunwald2004,Dawid2007}. As shorthand, $L(y,P) = L(y, a_P)$. Define the \emph{Bayes risk} by $\Lbar(P) := \inf_a \EE_{y \dist P}L(y, a) = \EE_{y \dist P}L(y, P)$. Similarly for any loss function define the \emph{regret} \cite{Grunwald2004}
$$
D_L(P,Q) := \EE_{y \dist P} L(y,Q) - \EE_{y \dist P} L(y, P).
$$ 
The regret measures how suboptimal the best action for distribution $Q$ is when played against distribution $P$. It should be obvious that the regret is always positive and equal to zero if $P=Q$.

In supervised learning, one assumes a fixed but unknown distribution $P_{XY} \in \mathcal{P}(X \times Y)$ over instance labels pairs. One wishes to find a function $f \in A^X$ that chooses a suitable action upon observing a given instance. Ideally $f$ should minimize the \emph{risk} $R_L(P_{XY}, f) := \EE_{P_{XY}} L(y,f(x))$. If we allow \emph{randomized} functions, ie Markov kernels $P_{A|X} : X \rightarrow \mathcal{P}(A)$ then we can extend the definition of risk to $R_L(P_{XY}, P_{A|X}) := \EE_{P_{XY}} \EE_{P_{A|X}} L(y,a)$. For the purpose of finding minimum risks, randomization does not help. Denote by
$$
\Rbar_L(P_{XY}) = \inf_{f \in A^X} R_L(P_{XY}, f)
$$
$$
f_{P_{XY}} = \arginf_{f \in A^X} R_L(P_{XY}, f)
$$
the minimum risk and Bayes optimal respectively. By standard manipulations 
$$
\Rbar_L(P_{XY})=\EE_{x \dist P_X}\Lbar(P_{Y|x}) = \EE_{(x, y) \dist P_{XY}} L(y, P_{Y|x}) 
$$ 
and $f_{P_{XY}}(x) = \arginf_a \EE_{P_{Y|x}} L(y,a)
$, where $P_{Y|X}$ is the Markov kernel obtained from applying Bayes rule to $P_{Y} \otimes P_{X|Y}$. In practice one is normally restricted to $f$ in some function class and only has a sample of $n$ iid draws from $P_{XY}$ with which to learn from. As our focus here is on ``preserving the information'' in $P_{XY}$, we shall in large part avoid such concerns.

\section{Supervised Feature Learning/ Loss and Experiment Specific Features}

For a multitude of reasons including but not limited to, computation, storage, the curse of dimensionality, increased classification performance, knowledge discovery and so on we may wish to process the instances through a (possibly randomized) feature map $P_{Z|X}$. For a given feature map, learning follows the protocol: First, nature draws $(x,y) \dist P_{XY}$. Second, the learner observes $z \dist P_{Z|x}$ and chooses an action $a$. Finally, the learner incurs loss $L(y,a)$. Diagrammatically,
$$
\xymatrix{
data = (x,y) \dist P_{XY} \ar[r] & \boxed{P_{Z|X}}  \ar[rr]^{(z,y)} && \boxed{f} \ar[rr]^{(f(z),y)} && L(f(z),y).
}
$$
By using the feature map we move from $P_{XY}$ to $P_{ZY}$ with
$$
\EE_{P_{ZY}} f = \EE_{(x, y) \dist P_{XY}} \EE_{z \dist P_{Z|x}} f(z,y)
$$
for all measurable $f: Z \times Y \rightarrow \RR $. Hence $P_{ZY} = P_Y \otimes (P_{Z|X} \circ P_{X|Y})$. Ideally $P_{ZY}$ should contain just as much ``information'' as $P_{XY}$, in sense that the \emph{feature gap}
$$
\Delta \Rbar_L(P_{XY}, P_{Z|X}) := \Rbar_L(P_{ZY}) - \Rbar_L(P_{XY})
$$
should be small. To be clear, $\Rbar_L(P_{ZY}) = \inf_{f \in A^Z} R_L(P_{ZY}, f) $, i.e. we are restricted to functions that \emph{only use the features}.

\begin{theorem}
For all joint distributions $P_{XY}$, feature maps $P_{Z|X}$ and loss functions $L$
$$
\Delta \Rbar_L(P_{XY}, P_{Z|X}) = \EE_{(x, z) \dist P_{XZ}} D_L(P_{Y|x} ,P_{Y|z})
$$
\end{theorem}

For proof see additional material. Hence the feature gap is the average regret suffered in using features versus raw instances \emph{when acting optimally for both}. In particular this means the feature gap is always non-negative.

\subsection{Link to Sufficiency and Conditional Independence}

The feature gap is closely related to the statistical notions of sufficiency and conditional independence. In particular we have the following theorem

\begin{theorem}[Blackwell-Sherman-Stein \cite{Torgersen1991}]
$\Delta \Rbar_L(P_{XY}, P_{Z|X}) = 0 $ for all loss functions if an only if $X$ and $Y$ are conditionally independent given $Z$.
\end{theorem}
In fact the Blackwell-Sherman-Stein theorem is even stronger, if we are also allowed to change the prior $P_Y$ on labels as well as the loss, and the feature gap remains zero then $Z$ is \emph{sufficient} for $X$ \cite{Torgersen1991,Le1964,Morse1966}. $Z$ contains all the useful information in $X$ for predicting $Y$, in both the average risk and minimax sense. If $Y$ is finite, then the vector of likelihood ratios $(\frac{dP_{X|y_1}}{dP_X}, \dots, \frac{dP_{X|y_n}}{dP_X} ) \in \RR^{|Y|}$ is always a sufficient for $X$. This in turn means the Bayesian posterior distribution is also sufficient for priors that do not assign some $Y$ zero mass. In many cases we can do better and find $Z$ sufficient for $X$, or close to, with $Z$ of lower dimension than $|Y|$ or even for $Z$ finite.

This observation has led to several classes of algorithms for supervised feature learning. One picks a loss with the property that $D_L(P,Q) = 0$ iff $P=Q$ and then uses this loss as a surrogate for testing sufficiency by finding
$$
\arginf_{P_{Z|X}} \Delta \Rbar_L(P_{XY}, P_{Z|X}).
$$
Of course if $\inf_{P_{Z|X}} \Delta \Rbar_L(P_{XY}, P_{Z|X}) = 0$ for one of these surrogates then by theorem 1 and as regret is always non negative, the feature gap will be zero for all losses. Some common surrogates include log loss leading to $D_L(P,Q) = D_{KL}(P,Q)$ which leads to the information bottleneck \cite{Tishby2000}. More general Bregman divergences lead to clustering with Bregman divergences \cite{Banerjee2005}. Finally, kernel mean based losses $L:T \times \mathcal{H} \rightarrow \RR$ with $\mathcal{H}$ a Hilbert space can also be used. Taking $\phi: Y \rightarrow \mathcal{H}$ and $L(y,v) = \lVert \phi(y) - v \rVert_{\mathcal{H}}^2$ with $\phi$ \emph{characteristic} \cite{Fukumizu2011} yields another suitable surrogate \cite{Wang2010}. In this case $D_L(P,Q) = \lVert \mu_P - \mu_Q \rVert_{\mathcal{H}}^2$, the squared distance between the kernel means of $P$ and $Q$.

For all the previous cases, algorithms exist for performing the minimization. These include alternating algorithms much like the Blahut-Arimoto algorithm of rate distortion theory \cite{Cover2012} in the first two cases, with something a bit more involved in the third (although it is restricted to linear, deterministic feature maps).

In practice, one might not know the exact loss function to use. Hence care must be taken in choosing a suitable surrogate or set of surrogates. We show in the examples section that the loss function can greatly influence how we rank features. This should be no of no surprise as the loss function \emph{defines} the relevant information contained in $P_{XY}$ \cite{Reid2009b}.

\subsection{Link to Deficiency}

If the loss is not known one can perform a worst case analysis
$$
\sup_{L, \lVert L \rVert \leq 1} \Delta \Rbar_L(P_{XY}, P_{Z|X}).
$$
Worse case differences in risk as the loss is varied have been studied extensively in the sub field of theoretical statistics known as the comparison of statistical experiments \cite{Torgersen1991,Le1964}. In this area the focus is placed on the \emph{experiments} $P_{X|Y}$ and $P_{Z|Y}$.
\begin{definition}
Let $P_{X|Y}$ and $P_{Z|Y}$ be experiments on $Y$, and $P_Y$ a distribution on $Y$. The \emph{weighted directed deficiency} from $P_{Z|Y}$ to $P_{X|Y}$ is equal to 
$$
\delta_{P_Y}(P_{Z|Y}, P_{X|Y}) = \inf_{P_{X|Z}} \EE_{y \dist P_Y} \lVert P_{X|y} - P_{X|Z} \circ P_{Z|y} \rVert
$$
\end{definition}
The weighted directed deficiency measures how close we can make $P_{Z|Y}$ to $P_{X|Y}$ in the sense of variational divergence by adding extra noise $P_{X|Z}$. It is closely related to approximate notions of sufficiency \cite{Le1964,Torgersen1991}, and it appears in an approximate version of the Blackwell-Sherman-Stein theorem.

\begin{theorem}[Randomization \cite{Torgersen1991}] For all $P_Y \in \mathcal{P}(Y)$ and for all experiments $P_{X|Y}$ and $P_{Z|Y}$,
$\Delta \Rbar_L(P_{XY}, P_{Z|X}) \leq \epsilon \lVert L \rVert$ if and only if $\delta_{P_Y}(P_{Z|Y}, P_{X|Y}) \leq \epsilon$
\end{theorem}
This theorem suggests a means to construct features when the loss function is not known, by minimizing the weighted directed deficiency. While this may appear difficult, one can exploit properties of the variational divergence that make calculating the weighted directed deficiency a $L_1$ minimization problem (see additional material). As long as the sets $X,Y$ and $Z$ are finite, fast methods exist to solve this problem. One can obtain features by finding
$$
\inf_{P_{Z|X}, \hat{P}_{X|Z}} \EE_{y \dist P_Y} \lVert P_{X|y} - \hat{P}_{X|Z} \circ P_{Z|X} \circ P_{X|y} \rVert
$$
and then using $P_{Z|X}$ as the feature map. This can be solved approximately through an alternating scheme of $L_1$ minimization problems (see additional material). Examples of how this method behaves on some toy problems are given in the examples section.

\section{Unsupervised Feature Learning}

One major drawback of the previous supervised feature learning methods is that they require some knowledge of $P_{Y|X}$ or $P_{X|Y}$. The first three methods also require some knowledge of the loss function of interest. These methods consider a single supervised task in isolation. They extract the information in $X$ that is relative to predicting $Y$. In many problems of interest we have access to a large data set of unlabelled samples drawn from $P_X$, however we may have limited knowledge of the tasks that $X$ will be \emph{used for}. We desire a feature map that provides a compact representation of $X$, \emph{that looses no information} about $X$. While at first this might seem vacuous, for example one could always just use the identity function, in many cases we can do much better. The data sets we tend to deal with have certain structure that we have not cared to directly specify in our models. This automated search for structure is what is behind the current deep learning fashion. 

Here we make the assumption that we have enough data to form an accurate estimate of $P_X$, the marginal distribution over instances, and ask the following question. Under what conditions can we guarantee that a feature map $P_{Z|X}$ does not lose more than $\epsilon$ information about $Y$ no matter what the relation between $X$ and $Y$ or the loss function? The only restriction we place on possible relationships $P_{XY}$ between $X$ and $Y$ is that the marginal distribution over instances is consistent with the one we have learnt. 
\begin{theorem}
For all feature maps $P_{Z|X}$, $\Delta \Rbar_L(P_{XY}, P_{Z|X}) \leq \epsilon \lVert L \rVert$ for all $P_{XY}$, label spaces $Y$ and loss functions $L$ if and only if there exists a $\hat{P}_{X|Z}$ such that $\EE_{x \dist P_X} \EE_{x' \dist \hat{P}_{X|Z} \circ P_{Z|x}} \mathbb{1} (x' \neq x) \leq \epsilon$
\end{theorem}
In order to minimize the information lost from $X$, one needs to be able to reconstruct $X$ from $Z$ with high probability. We show in the next section that under some of the heuristic justifications of deep learning techniques like the autoencoder \cite{Vincent2008} and the deep belief network \cite{Hinton2006}, one is solving a surrogate to this problem.

Theorem 4 also highlights the connection between feature learning and \emph{reconstruction}. Reconstructing well is equivalent to finding generically good features. Theorem 4 also makes no use of interesting structure of the instance space $X$, effectively using the discrete metric on $X$, $d(x,x') = 1$ if $x \neq x'$. If one makes a smoothness assumption on the experiments of interest, a different version of theorem 4 is obtained.
\begin{definition}
For all joint distributions $P_{XY}$ and losses $L$ the \emph{reconstruction regret} is given by 
$$
D_{r}(x,x') = D_L(P_{Y|x}, P_{Y|x'}) = \EE_{y \dist P_{Y|x}} L(y, f_{P_{XY}}(x')) - \EE_{y \dist P_{Y|x}} L(y, f_{P_{XY}}(x))
$$
\end{definition}
The reconstruction regret is the regret suffered in choosing actions based on a nearby $x'$ when in fact one should have used $x$. If we assume that $X$ is equipped with a metric $d: X \times X \rightarrow \RR$, then we might wish to reconstruct well with respect to this metric.
\begin{theorem}
For all feature maps $P_{Z|X}$ the following are equivalent
\begin{enumerate}
	\item $\exists \hat{P}_{X|Z}$ such that $\EE_{x \dist P_X} \EE_{x' \dist \hat{P}_{X|Z} \circ P_{Z|x} } d(x,x') \leq \epsilon$
	\item For all distributions $P_{XY}$ and loss functions $L$ with $D_r(x,x') \leq \lambda d(x,x')\ \forall x, x'$, \\	$\Delta \Rbar_L(P_{XY}, P_{Z|Y}) \leq \epsilon \lambda$
\end{enumerate}
\end{theorem}
For proof see additional material. Theorem 4 follows by taking $d$ to be the discrete metric on $X$.

\subsection{Surrogates Approaches Motivated by Theorem 4}

Theorem 4 requires one to be able to reconstruct $X$ from the features $Z$ with high probability if one wishes generically good features. There are many surrogates to this problem. Many existing feature learning methods are motivated through an appeal to the Infomax principle \cite{Linsker1989}. Features should be chosen to maximize the mutual information $I(X;Z)$ or equivalently to minimize the conditional entropy $H(X|Z)$.
\begin{theorem}[Hellman-Raviv \cite{Hellman1970}]
Let $X$ and $Z$ be finite spaces. For all feature maps $P_{Z|X}$ and priors $P_X$,
$$
\inf_{\hat{P}_{X|Z}} \EE_{x \dist P_X} \EE_{x' \dist \hat{P}_{X|Z} \circ P_{Z|x}} \mathbb{1} (x' = x) \leq \frac{1}{2} H(X|Z).
$$
\end{theorem}
Hence the conditional entropy bounds the smallest probability of error possible when one attempts to reconstruct $X$ from the feature map $P_{Z|X}$. One can view the Infomax principle as being a surrogate to reconstruction error. By exploiting various representations of $H(X|Z)$, many other surrogates to reconstructing with high probability can be obtained \cite{Bengio2013,Vincent2008}. For example, by properties of the KL divergence
$$
H(X|Z) = \EE_{(x,z) \dist P_{XZ}} -\log(P_{X|z}(x)) = \inf_{\hat{P}_{X|Z}}  \EE_{(x,z) \dist P_{XZ}} -\log(\hat{P}_{X|z}(x)).
$$
If we restrict the possible $\hat{P}_{X|Z}$ to distributions of the form $P_{X|z} = \mathcal{N}(f(z), \sigma^2)$ (normal distributions with mean $f(z)$) for some function $f: Z \rightarrow X$ and standard deviation $\sigma$, we obtain
$$
H(X|Z) \leq \inf_{f,\sigma}  \EE_{(x,z) \dist P_{XZ}} \frac{1}{2 \sigma ^ 2}(x - f(z))^2 + \log(\sqrt{2 \pi} \sigma).
$$
If we restrict the possible feature maps to $P_{Z|x} = \delta_{g(x)}$ then we the autoencoder. Hence the autoencoder can be seen as a surrogate to theorem 4. Its use can also be justified by theorem 5. Many other feature learning methods such as K-means and principle component analysis can be seen as specific instances of the autoencoder, $g$ are linear projections for PCA and $Z$ is finite for K-means.

\subsection{Rate Distortion Theory}

Rate-distortion theory provides lower bounds on the \emph{distortion}, or in our terminology $\Rbar_L(P_{ZY})$, in terms of the \emph{rate} $I(X;Z)$ of the form $\phi^{-1}_L(I(X;Z)) \leq \Rbar_L(P_{ZY}) $ with $\phi_L$ the \emph{rate distortion function}. 
$$
\phi_L(d)= \inf_{P_{A|Y}, \EE_{P_{YA}} L \leq d} I(Y;A). 
$$
Determining this function involves solving a series of convex problems, for which a fast iterative algorithm exists \cite{Cover2012}. The end to end performance of the complete system is captured in the rate distortion function, the quality of the feature map by $I(X;Z)$. This bound provides a ranking of feature maps that depends only on the loss of interest and the mutual information of the feature map, and more importantly \emph{not on the experiment}. Combined with theorem 5 one obtains bounds of the form
$$
\phi^{-1}_L(I(X;Z)) \leq \Rbar_L(P_{ZY}) \leq \Rbar_L(P_{XY}) + \frac{1}{2}H(X|Z) \lVert L \rVert , \ \forall L.
$$
Are there better surrogates? Ideally we wish to calculate $\Rbar_L(P_{ZY})$, however this requires knowledge of $P_{X|Y}$ and not just the feature map $P_{Z|X}$ and marginal $P_X$. We can calculate $I(X;Y)$ and rely on rate distortion and deficiency theory to provide bounds. This begs the question, are there better surrogates? If we know the loss function can we do better than mutual information for providing performance bounds? At least in the case of the lower bound the answer is yes. In \cite{2573956}, a large class of generalized information measures are considered. For each of these information measures a rate-distortion theorem is obtained and in many cases using one of these instead of mutual information provides \emph{tighter} lower bounds. 
\begin{definition}
For convex $f: \RR^+ \rightarrow \RR$ with $f(1) = 0$, the \emph{$f$-information} of a joint distribution $P_{XY}$ is given by
$$
I_f(P_{XY}) = \EE_{P_{XY}} f(\frac{d(P_X \otimes P_Y)}{d P_{XY}}).
$$
\end{definition}
We present in the illustrations section an example of when using one of these measures of information provides a tighter bound than mutual information. This observation may have algorithmic implications. Ultimately the feature map $P_{Z|X}$ will be restricted to lie in some function class. If $L$ is known it may be better to optimize one of these general forms of information rather than mutual information.

\subsection{Hierarchical Learning of Features}

One of the main tenets of the deep learning paradigm is that features should be learnt in a hierarchical fashion. Rather than learning a single feature map, one learns a chain
$$
\xymatrix{
X = Z_0 \ar@/^/[rr]^{P_{Z_1|Z_0}} && \ar@/^/[ll]^{\hat{P}_{Z_0|Z_1}} Z_1 \ar@/^/[rr]^{P_{Z_2|Z_1}} && \ar@/^/[ll]^{\hat{P}_{Z_1|Z_2}} Z_2 \ar@/^/[rr]^{P_{Z_3|Z_2}} && \ar@/^/[ll]^{\hat{P}_{Z_2|Z_3}} \dots \ar@/^/[rr]^{P_{Z_n|Z_{n-1}}} && \ar@/^/[ll]^{\hat{P}_{Z_{n-1}|Z_n}} Z_n
}
$$
with final feature map $P_{Z_n |X} = P_{Z_n|Z_{n-1}} \circ \dots \circ P_{Z_1|X}$ the composition of all the feature maps in the chain, and final reconstruction given by $\hat{P}_{X|Z_n} = \hat{P}_{X|Z_1} \circ \dots \circ \hat{P}_{Z_{n-1}|Z_n}$. Such a scheme has obvious computational advantages, one can learn each layer in a greedy fashion. To analyse the entire system, one can invoke a union bound obtaining
$$
\EE_{x \dist P_X} \EE_{x'\dist \hat{P}_{X|Z_n} \circ P_{Z_n|x}} \mathbb{1}(x \neq x') \leq \sum\limits_{i=0}^{n-1} \EE_{z_i \dist P_{Z_i}} \EE_{z_i'\dist \hat{P}_{Z_i|Z_{i+1}} \circ P_{Z_{i+1}|Z_i}} \mathbb{1}(z_i \neq z_i')
$$
i.e., the probability of reconstruction error for the entire system is bounded by the sum of the probability of reconstruction errors for each layer. See additional material for a proof. Hence the deep belief network and other hierarchical methods can be seen as solving a surrogate to theorem 4. 

\subsection{Semi Supervised Learning and Transfer of Generalization Bounds}

In semi supervised learning one wishes to learn a classifier $f \in A^X$ from a data set comprising of $n$ draws from $P_{XY}$ and $m$ draws from $P_X$, where normally $m >> n$. To tackle this problem one can learn a \emph{representation} of $X$ via a feature map $P_{Z|X}$ from the unlabelled data. One can then learn a classifier $g \in A^Z$ from the labelled data $(z_i, y_i) \dist P_{ZY}$, $z_i \dist P_{Z|x_i}$. Theorem 5 allows one to analyse the generalization performance of such a joint system. If something is known about the sample complexity of learning $g$ and of learning $P_{Z|X}$ then theorem 5 allows one to combine these to give a sample complexity for learning both. Much is known about the sample complexity of supervised learning. For the sample complexity of (some) reconstruction schemes we point the reader to the recent work \cite{Maurer2008,Biau2008}. These works give sample complexity bounds for many different reconstruction schemes under square loss, in particular k-means, principle component analysis and sparse coding. Our results allow one to \emph{transfer} these results to the semi supervised learning domain.

\section{Illustrations}
In this section we give some \emph{simple} examples of how the different feature learning schemes discussed operate in practice. We also give examples of when one can learn sufficient features for a particular experiment as well as when it is possible to learn generic features.

\textbf{Experiment Specific Features.} Let $Y = \RR$ with $X = \RR^n$ and $P_{X|Y}$ given by the product of $n$ normal distributions with mean $y$ and variance $1$. It is easy to verify that the sample mean $\phi: X \rightarrow \RR$ is a sufficient statistic meaning that at least for this experiment we can greatly compress the information contained in $X$. However, if we take as a prior for $Y$ a normal distribution of mean $0$ and variance $1$, then the marginal distribution $P_X$ will not be concentrated on a set of smaller dimension nor have any particularly interesting structure. Hence we can not find interesting generic features in this case.

\textbf{Experiment and Loss Specific Features.} Let $Y = \{-1,1\}$ with $P_{X|y}=\mathcal{N}(y,1)$. For this experiment, 0-1 loss ($L_{01}$) and a uniform prior the Bayes optimal $f$ is given by $f(x) = 1$ if $x > 0$ as $P(-1|x)>\frac{1}{2}$ and $f(x) = -1$ otherwise as $P(-1|x)\leq\frac{1}{2}$ . It is easy to show that $\Delta \Rbar_{L_{01}}(P_{XY},f) = 0$, all we need is the output of $f$. However if we change the loss to a cost sensitive loss $L_c$ \cite{Reid2009b} where misclassifying a $1$ is more costly than a $-1$, we no longer have $\Delta \Rbar_{L_c}(P_{XY},f) = 0$, as this would change the optimal threshold for classifying a $1$ versus $-1$. However, if there was a jump discontinuity in $P(-1|X)$, ie it jumped from say $0.4 $ to $0.6$ as $x$ crossed over $x = 0$ then the feature gap would be zero for a broader range of cost sensitive losses. Once again there are not generic features of interest.

\textbf{Loss Sensitive versus Loss Insensitive Features.} Let $Y = \{1,2,3\}$ with a uniform prior for $Y$ and $P_{X|Y}$ given by the normal distributions in the figure below. Consider the feature space $Z = \{1,2\}$. Below are plots of the features learnt by two different feature learning schemes. The first is the loss insensitive weighted directed deficiency minimization method. The second is the information bottleneck \emph{where we know before hand} that misclassifying a $2$ is more costly than misclassifying one of the others. A loss of this form is achieved by tilting the standard brier loss \cite{Reid2009b} toward class $2$. The green regions are those $x$ that are mapped to the feature $1$, the blue are those mapped to $2$. 

\begin{figure}[ht]
\centering
\includegraphics[width=1.1\linewidth]{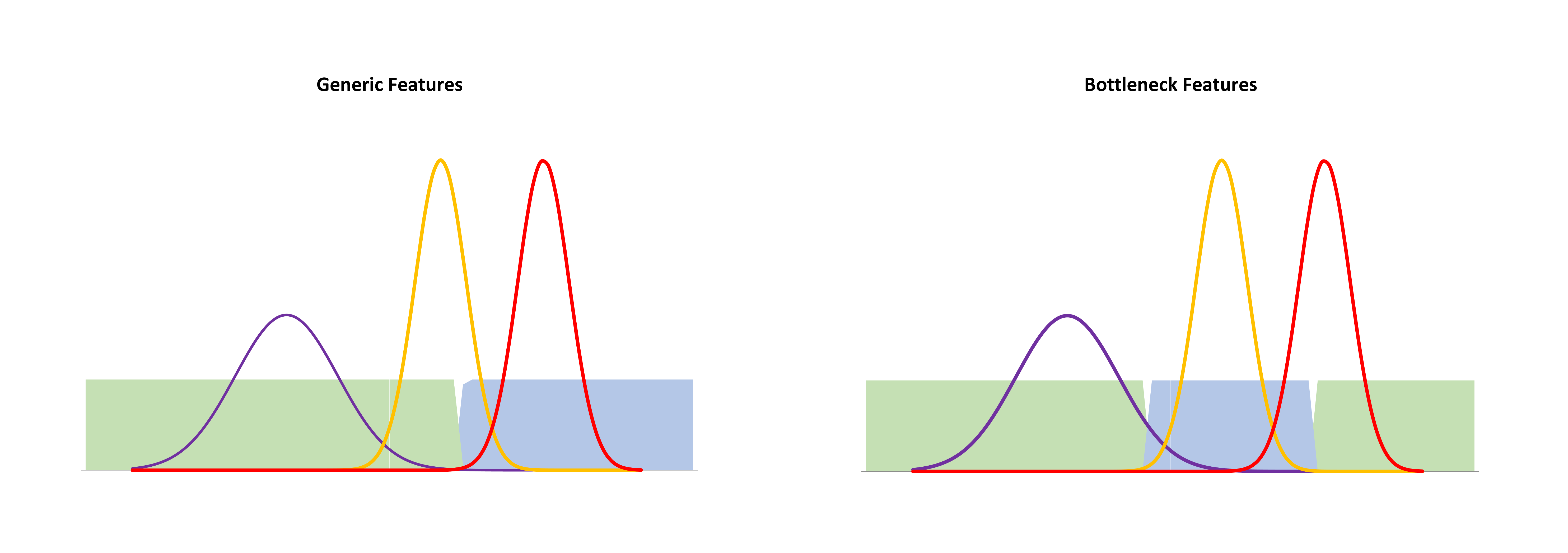}
\caption{Loss Sensitive versus Loss Insensitive Features, see text}
\end{figure}

We can see even in this simple example that the loss function matters when determining sensible features. While the weighted directed deficiency method divides $X$ into regions that allow good reconstruction of all the class conditionals, the bottleneck features \emph{focus} on separating class $2$ as dictated by the loss function. While the weighted directed deficiency $\delta_{P_Y} (P_{X|Y}, P_{Z|Y})$ being $0.629$ and $0.698$ respectively indicating that from a \emph{worst case} perspective the two feature maps are very similar. However, for the particular loss we have used the \emph{feature gap} is very different, $1.075$ versus $0.325$.

\textbf{Learning Generic Features.} All previous examples have considered a \emph{fixed} experiment. When learning features in an unsupervised fashion, one wishes to find features that work for all experiments that use $X$. There are many examples of when this is possible, and they all boil down to some sort of manifold assumption. If $P_X$ is concentrated on some lower dimensional subset of $X$, then one can find generic features.

\textbf{Rate Distortion Lower Bounds.}
As an example of the different bounds one can obtain using $f$-informations, we consider a simple example where $Y=\{0,1\}$ and the loss is a cost sensitive misclassification loss with $L(0,1) = 1$ and $L(1,0) = 4$. We consider the feature map 
$$
P_{Z|X} = \left( \begin{matrix}
0.8 & 0.1 & 0.1 \\
0.1 & 0.4 & 0.5
\end{matrix} \right)
$$
given as a row stochastic matrix with uniform prior $P_X$. We consider $f(x) = (\sqrt{x} - 1)^2$ resulting in Hellinger information. Below are plots of the rate distortion curves for both mutual information (red) and Hellinger information (blue) as well as the informations of the channel (the dashed horizontal lines). The black vertical line represents the lower bound on the distortion. For this channel Hellinger information gives a tighter lower bound.
\begin{figure}[th]
\centering
\includegraphics[width=0.73\linewidth]{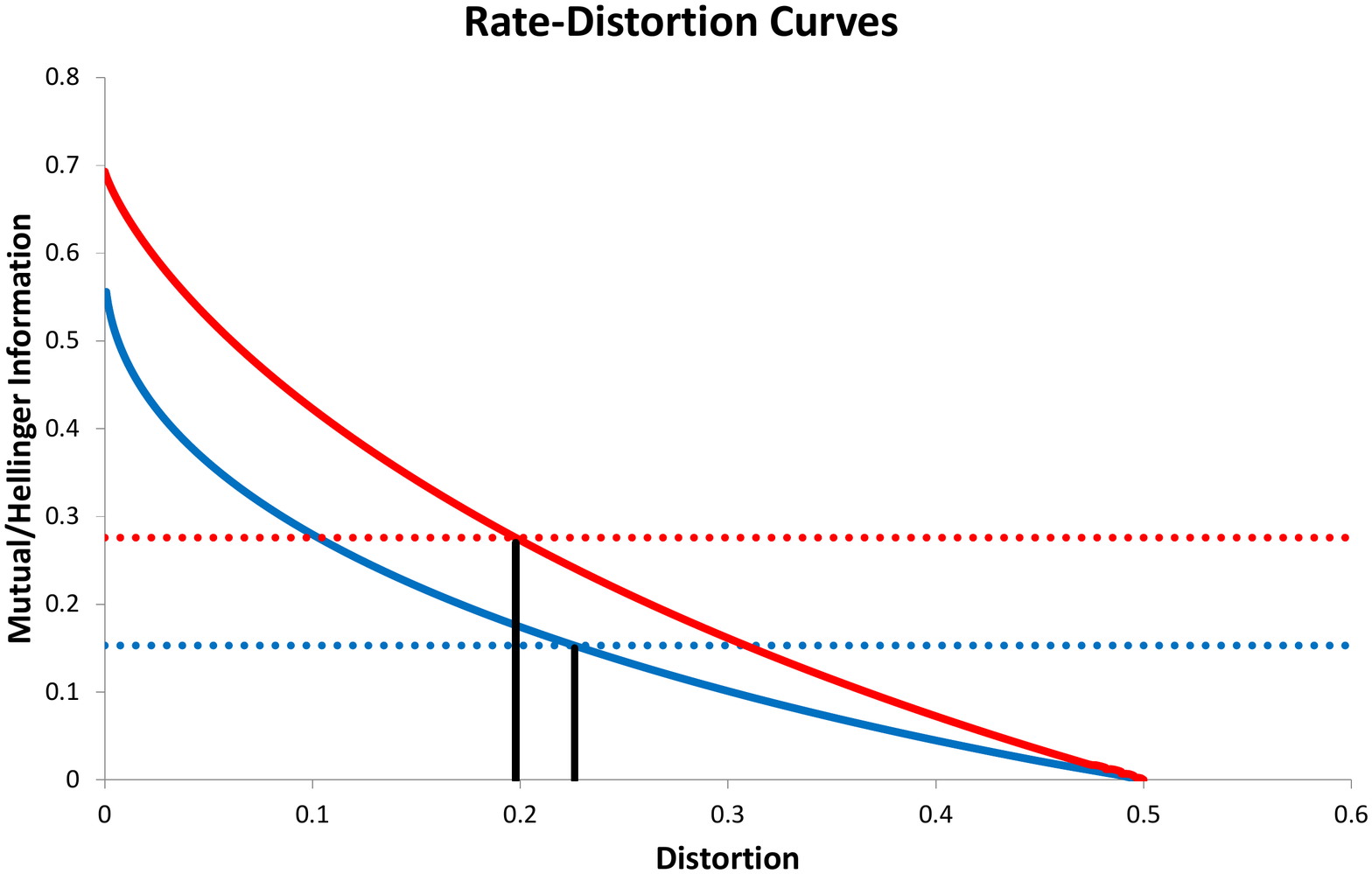}
\caption{Generalized Rate-Distortion Plots, see text}
\end{figure}
For further illustrations see additional material.

\section{Conclusion}

Automated feature learning methods have produced remarkable empirical results, however little theory exists explaining their performance. This paper provides direction as to how progress the theory. To this end, we have placed several current supervised feature learning methods in a general framework, provided a novel loss insensitive method for learning features as well as providing novel means of transferring regret bounds from unsupervised feature learning methods to supervised learning methods. Finally, we have shown the usefulness of rate-distortion theory and its under utilized generalizations in ascertaining the quality of learnt features.

\newpage
\small
\bibliographystyle{plain}
\bibliography{ref}

\newpage

\section{Additional Material}

\subsection{Background on Proper Losses}

Here we review some material that greatly eases working with proper loss functions and highlights the connection between loss, Bayes risk, regret and Bregman Divergences \cite{Grunwald2004,Dawid2007}.

\begin{definition}
A loss function $L: Y \times \mathcal{P}(Y) \rightarrow \RR$ is \emph{proper} if for all $P \in \mathcal{P}(Y)$
$$
P \in \arginf_{Q \in \mathcal{P}(Y)} \EE_{y \dist P} L(y,Q) 
$$
\end{definition}

Any loss function can be \emph{properized}.

\begin{theorem}
Let $L: Y \times A \rightarrow \RR$ be a loss. For $P\in \mathcal{P}(Y)$ Define 
$$
a_P = \arginf_a \EE_{y \dist P} L(y,a)
$$
where we arbitrarily pick an $a \in \arginf_a \EE_{y \dist P} L(y,a)$ if there are multiple. Then $\hat{L}(y,P) = L(y,a_P)$ is proper.
\end{theorem}
It is possible that by using this trick we remove useful actions $a\in A$. However, for the purpose of calculating expected risks we do not require these actions. From $L$, one can define a \emph{regret}
$$
D(P,Q) = \EE_{y \dist P} L(y,a_Q) - \EE_{y \dist P} L(y,a_P)
$$
which measures how suboptimal the best action for the distribution $Q$ is when played against the distribution $P$. One does not need knowledge of the original loss to construct $\hat{L}$, only the \emph{Bayes risk}  
$$
\Lbar(P) = \inf_a \EE_{y \dist P} L(y,a)
$$
is needed. From this one can reconstruct $\hat{L}$, and hence $L$ for the purposes of calculating minimum expected risks. This is achieved by taking the 1-homogeneous extension of $\Lbar$ 
\begin{align*}
\tilde{\Lbar}: \RR_+^{|Y|} &\rightarrow \RR \\
v &\mapsto \lVert v \rVert_1 \Lbar(\frac{v}{\lVert v \rVert_1})
\end{align*}
and differentiating/taking super gradients. The following three theorems highlight the usefulness of the 1-homogeneous extension.
\begin{definition}{Super Gradient Function}
Let $f: C \subseteq \RR^n \rightarrow \RR$ be a concave function. Then 
$$
\nabla f: C \rightarrow \RR^n
$$ 
is a super gradient function if for all $x\in C$, $\nabla f(x) \in \hat{\partial}f(x)$
\end{definition}

\begin{theorem}
For any concave 1-homogeneous function $f: C \subseteq \RR^n \rightarrow \RR$ and any super-gradient function $\nabla f$, 
$$
f(x) = \langle x, \nabla f (x) \rangle
$$ 
\end{theorem}

\begin{theorem}
For any concave $\Lbar: \mathcal{P}(Y) \rightarrow \RR$
$$
L(y,P) = \langle \delta_y , \nabla \tilde{\Lbar}(P) \rangle
$$
is a proper loss.
\end{theorem}

\begin{theorem} The regret derived from a proper loss $L$ is equal to the \emph{Bregman divergence} defined by $\tilde{\Lbar}(P)$.
$$
D(P,Q) = D_{\Lbar}(P,Q) = D_{\tilde{\Lbar}}(P,Q) = \tilde{\Lbar}(Q) + \langle  P - Q  , \nabla \tilde{\Lbar}(Q) \rangle - \tilde{\Lbar}(P) = \langle P, \nabla \tilde{\Lbar}(Q) - \nabla \tilde{\Lbar}(P) \rangle.
$$
\end{theorem}

Finally give a concave 1-homogeneous Bayes risk $\tilde{\Lbar}$ and a vector $\pi \in \RR_+^{|Y|}$ one can \emph{tilt} $\tilde{\Lbar}$ by $\pi$ yielding
\begin{align*}
\tilde{\Lbar}_{\pi} : \RR_+^{|Y|} &\rightarrow \RR \\
\tilde{\Lbar}_{\pi}(v) &= \langle \pi, v \rangle \tilde{\Lbar}(\frac{\pi v}{\langle \pi, v \rangle})
\end{align*}
with $\pi v$ the element wise product of $\pi$ and $v$. It is easily verified that this new function is both concave and 1-homogeneous. The tilting has the effect of making certain elements of $Y$ being more important in the resulting loss. For example if we start with a symmetric Bayes risk like Shannon entropy, and tilt by the vector $(1, 10, 1)$, then the resulting loss places more importance on predicting $y_2$ correctly. This is analogous to how cost-sensitive misclassification losses are produced from 01 loss.

\subsection{Background on the Information Bottleneck}

For a given joint distribution $P_{XY}$ and loss function $L$, the information bottleneck/ clustering with Bregman divergences attempt to extract a feature map by solving 
$$
\inf_{P_{Z|X}} \beta \Delta \Rbar_L(P_{XY}, P_{Z|X}) + I(X;Z)
$$
i.e. a regularized feature gap, with the mutual information $I(X;Z)$ serving as the regularizer. This can be solved by an alternating algorithm. Here we review the derivation of this algorithm.

\begin{theorem}
\begin{align*}
&\inf_{P_{Z|X}} \beta \Delta \Rbar_L(P_{XY}, P_{Z|X}) + I(X;Z) \\
= &\inf_{P_{Z|X}} \inf_{\hat{P}_{\Theta|Z}} \inf_{\hat{P}_{Z}} \beta \EE_{x \dist P_{X}} \EE_{z\dist P_{Z|x}} D_{\Lbar}(P_{\Theta|x},\hat{P}_{\Theta|z}) + \EE_{x \dist P_{X}} D_{KL}(P_{Z|x},\hat{P}_{Z})
\end{align*}
\end{theorem}

For the proof we require the following lemma from \cite{Banerjee2005}

\begin{lemma}
For all concave $\Lbar: X \rightarrow \RR$ and distributions $P\in \mathcal{P}(X)$ 
$$
\EE_{P} X \in \arginf_{x} \EE_{y \dist P} D_{\Lbar}(y,x).
$$
The mean is the expected Bregman divergence minimizer.
\end{lemma}
We can now prove the theorem
\begin{proof}
Firstly,
$$
I(X;Y) = \EE_{x \dist P_{X}} D_{KL}(P_{Z|x},P_{Z}) = \inf_{\hat{P}_{Z}} \EE_{x \dist P_{X}} D_{KL}(P_{Z|x},\hat{P}_{Z}).
$$
as $\EE_{x\dist P_{X}} P_{Z|x} = P_{Z}$. Secondly 
\begin{align*}
\EE_{x \dist P_{X}} \EE_{z\dist P_{Z|x}} D_{\Lbar}(P_{\Theta|x},P_{\Theta|z}) &= \EE_{Z \dist P_{Z}} \EE_{x\dist P_{X|z}} D_{\Lbar}(P_{\Theta|x},P_{\Theta|z}) \\
&= \EE_{Z \dist P_{Z}} \inf_{\hat{P}_{\Theta|z}}\EE_{x\dist P_{X|z}} D_{\Lbar}(P_{\Theta|x},\hat{P}_{\Theta|z}) \\
&= \inf_{\hat{P}_{\Theta|Z}} \EE_{Z \dist P_{Z}} \EE_{x\dist P_{X|z}} D_{\Lbar}(P_{\Theta|x},\hat{P}_{\Theta|z})
\end{align*}
as $\EE_{x\dist P_{X|z}} P_{\Theta|x} = P_{\Theta|z}$. Combining gives 
\begin{align*}
&\inf_{P_{Z|X}} \EE_{x \dist P_{X}} \beta \EE_{z\dist P_{Z|x}} D_{\Lbar}(P_{\Theta|x},P_{\Theta|z}) + \EE_{x \dist P_{X}} D_{KL}(P_{Z|x},P_{Z}) \\
= & \inf_{P_{Z|X}} \inf_{\hat{P}_{\Theta|Z}} \inf_{\hat{P}_{Z}} \beta \EE_{x \dist P_{X}} \EE_{z\dist P_{Z|x}} D_{\Lbar}(P_{\Theta|x},\hat{P}_{\Theta|z}) + \EE_{x \dist P_{X}} D_{KL}(P_{Z|x},\hat{P}_{Z}).
\end{align*}
This completes the proof.
\end{proof}

The above theorem allows one to (at least approximately) find loss specific features.

\subsection{Background on Loss Insensitive Feature Learning}

Recall that loss insensitive feature learning seeks to find a feature map $P_{Z|X}$ and a re-constructor $\hat{P}_{X|Z}$ that minimize 
$$
\inf_{P_{Z|X}, \hat{P}_{X|Z}} \EE_{y \dist P_Y} \lVert P_{X|y} - \hat{P}_{X|Z} \circ P_{Z|X} \circ P_{X|y} \rVert.
$$
We show how this can be achieved by an alternating pair of linear programs. Assuming that $X,Y,Z$ are all finite sets, $P_{X|Y}, P_{Z|X}$ and $\hat{P}_{Z|X}$ can be represented by column stochastic matrices $T,F,R$ respectively, with composition represented as matrix multiplication. Furthermore $P_Y$ can be represented by a probability vector $v$. The variational divergence between two distributions is the $L_1$ distance between their probability vectors \cite{Reid2009b}. For fixed $F$ taking an infimum over $R$ means solving the following linear program
\begin{align*}
\inf_{Z_{ij}, R_{ij}} & \sum\limits_{i = 1}^{|X|} \sum\limits_{j = 1}^{|Y|} Z_{ij} \\
\text{subject to}\ & Z_{i,j}, R_{i,j} \geq 0\  \forall i, j \\
& \sum\limits_{i=1}^{|X|}R_{i,j} = 1 \ \forall j \\
& \lvert v_i T_{ij} -  v_i \sum\limits_{k = 1}^{|Z|} \sum\limits_{h = 1}^{|X|} R_{ik} F_{kh} T_{h j} \rvert \leq Z_{ij} \ \forall i, j.
\end{align*}
The final constraint can be written as a pair of linear constraints. Fixing R and taking an infimum over $F$ means solving the following 
\begin{align*}
\inf_{Z_{ij}, F_{ij}} & \sum\limits_{i = 1}^{|X|} \sum\limits_{j = 1}^{|Y|} Z_{ij} \\
\text{subject to}\ & Z_{i,j}, F_{i,j} \geq 0\  \forall i, j \\
& \sum\limits_{i=1}^{|Z|}F_{i,j} = 1 \ \forall j \\
& \lvert v_i T_{ij} -  v_i \sum\limits_{k = 1}^{|Z|} \sum\limits_{h = 1}^{|X|} R_{ik} F_{kh} T_{h j} \rvert \leq Z_{ij} \ \forall i, j.
\end{align*}
Alternating these two minimizations provides means to find loss insensitive features.

\subsection{Proofs for Some Theorems in Main Text}

\subsubsection{Proof of Theorem 1}

\begin{theorem*}
For all joint distributions $P_{XY}$ and feature maps $P_{Z|X}$  
$$
\Delta \Rbar_L(P_{XY}, P_{Z|X}) = \EE_{(x, z) \dist P_{XZ}} D_L(P_{Y|x} ,P_{Y|z})
$$
\end{theorem*}

\begin{proof}
\begin{align*}
\Rbar_L(P_{ZY}) - \Rbar_L(P_{XY}) &= \EE_{P_{ZY}} L(y, P_{Y|z})  - \EE_{P_{XY}} L(y, P_{Y|x}) \\
&= \EE_{(x, y) \dist P_{XY}} \EE_{z \dist P_{Z|x}} [L(y, P_{Y|z}) - L(y, P_{Y|x})] \\
&= \EE_{(x, z) \dist P_{XZ}} \EE_{y \dist P_{Y|x}} [L(y, P_{Y|z}) - L(y, P_{Y|x})] \\
&= \EE_{(x, z) \dist P_{XZ}} D_L(P_{Y|x} ,P_{Y|z})
\end{align*}
where the second last line follows from the fact that $Y \bot Z | X$ as $Y \rightarrow X \rightarrow Z$ forms a Markov chain.
\end{proof}

\subsubsection{Proof of Theorem 5}

\begin{theorem*}
For all feature maps $P_{Z|X}$ the following are equivalent
\begin{enumerate}
	\item $\exists \hat{P}_{X|Z}$ such that $\EE_{x \dist P_X} \EE_{x' \dist \hat{P}_{X|Z} \circ P_{Z|x} } d(x,x') \leq \epsilon$
	\item For all distributions $P_{XY}$ and loss functions $L$ with $D_r(x,x') \leq \lambda d(x,x')$, \\	$\Delta \Rbar_L(P_{XY}, P_{Z|Y}) \leq \epsilon \lambda$
\end{enumerate}
\end{theorem*}
\begin{proof}{[$1 \Rightarrow 2$]}
Let $f_{P_{XY}}$ be the Bayes optimal for $P_{XY}$ and $L$, and consider the following randomized function $P_{A|Z} = f_{P_{XY}} \circ \hat{P}_{X|Z}$, i.e. the composition of the Bayes optimal and the re-constructor.
\begin{align*}
\Delta \Rbar_L(P_{XY}, P_{Z|Y}) &\leq R_L(P_{ZY}, P_{A|Z}) - \Rbar_L (P_{XY}) \\
&= \EE_{(z, y) \dist P_{ZY}} \EE_{x' \dist \hat{P}_{X|z}} L(y,f_{P_{XY}}(x')) - \EE_{(x, y) \dist P_{XY}} L(y,f_{P_{XY}}(x)) \\
&= \EE_{(x, y) \dist P_{XY}} \EE_{x' \dist \hat{P}_{X|Z} \circ P_{Z|x}} [L(y,f_{P_{XY}}(x')) - L(y,f_{P_{XY}}(x))] \\
&= \EE_{(x, y) \dist P_{XY}} \EE_{x' \dist \hat{P}_{X|Z} \circ P_{Z|x}} [L(y,f_{P_{XY}}(x')) - L(y,f_{P_{XY}}(x))] \\
&= \EE_{x \dist P_{X}} \EE_{x' \dist \hat{P}_{X|Z}\circ P_{Z|x}} \EE_{y \dist P_{Y|x}} [L(y,f_{P_{XY}}(x')) - L(y,f_{P_{XY}}(x))] \\
&= \EE_{x \dist P_{X}} \EE_{x' \dist \hat{P}_{X|Z}\circ P_{Z|x}} D_r(x,x') \\
&\leq \EE_{x \dist P_{X}} \EE_{x' \dist \hat{P}_{X|Z}\circ P_{Z|x}} \lambda d(x,x') \\
&\leq \epsilon \lambda
\end{align*}

\end{proof}

\begin{proof}{[$2 \Rightarrow 1$]}
Let $Y=X$, $A=X$ and $L(x',x) = d(x',x)$. Finally let $P_{XY} = P_X \otimes id_X$, i.e. draw $x\dist P_X$ and return $(x,x)$. It is easy to confirm that $f_{P_{XY}}(x) = x$ and $\Rbar_L(P_{XY}) = 0$ and $D_r(x',x) = d(x',x)$. By $2$
\begin{align*}
\epsilon &\geq \Delta \Rbar_L(P_{XY}, P_{Z|Y}) \\
&= \Rbar_L(P_{ZY}) \\
&= \inf_{P_{A|X} \in \mathcal{P}(X')^Z} \EE_{P_{ZY}} \EE_{P_{A|X}} L(y,a) \\
&= \inf_{P_{A|Z} \in \mathcal{P}(X')^Z} \EE_{x \dist P_X}  \EE_{z \dist P_{Z|x}} \EE_{P_{X'|z}} d(x',x)
\end{align*}
hence $1$ is satisfied.
\end{proof}

\subsubsection{Hierarchical Learning of Features Proof}

\begin{theorem*}
For all chains of feature maps and reconstruction functions
$$
\xymatrix{
X = Z_0 \ar@/^/[rr]^{P_{Z_1|Z_0}} && \ar@/^/[ll]^{\hat{P}_{Z_0|Z_1}} Z_1 \ar@/^/[rr]^{P_{Z_2|Z_1}} && \ar@/^/[ll]^{\hat{P}_{Z_1|Z_2}} Z_2 \ar@/^/[rr]^{P_{Z_3|Z_2}} && \ar@/^/[ll]^{\hat{P}_{Z_2|Z_3}} \dots \ar@/^/[rr]^{P_{Z_n|Z_{n-1}}} && \ar@/^/[ll]^{\hat{P}_{Z_{n-1}|Z_n}} Z_n
}
$$
the probability of reconstruction error for the entire chain is bounded by the the sum of the reconstruction errors for each layer
$$
\EE_{x \dist P_X} \EE_{x'\dist \hat{P}_{X|Z_n} \circ P_{Z_n|x}} \mathbb{1}(x \neq x') \leq \sum\limits_{i=0}^{n-1} \EE_{z_i \dist P_{Z_i}} \EE_{z_i'\dist \hat{P}_{Z_i|Z_{i+1}} \circ P_{Z_{i+1}|Z_i}} \mathbb{1}(z_i \neq z_i')
$$
\end{theorem*}

\begin{proof}
Let $(z_0, z_1 ,\dots, z_n)$ be the ``true" elements at each level of the chain and $(z'_0, z'_1, \dots , z'_{n-1})$ their reconstructions. Consider the joint distribution $\boldsymbol{P}$ with
\begin{align*}
\boldsymbol{P}(z_0, z_1 ,\dots, z_n, z'_0, z'_1, \dots , z'_{n-1}) &=P(z_0) P(z_1| z_0)P(z_2|z_1)\dots P(z_n| z_n-1) P(z'_{n-1}| z_n)\dots P(z'_0| z'_1) \\
&= P_X(z_0) P_{Z_1|z_0}(z_1) \dots P_{Z_n| z_{n-1}}(z_n) \hat{P}_{Z_{n-1}| z_n}(z'_{n-1}) \dots \hat{P}_{Z_0|z'_1}(z'_0).
\end{align*}
Under this joint distribution
\begin{align*}
\boldsymbol{P}(z_0 \neq z'_0) &= \boldsymbol{P}(z_0 \neq z'_0 \cap z_1 = z'_1) + \boldsymbol{P}(z_0 \neq z'_0 \cap z_1 \neq z'_1) \\
&\leq \boldsymbol{P}(z_0 \neq z'_0 \cap z_1 = z'_1) + \boldsymbol{P}(z_1 \neq z'_1) 
\end{align*}
To complete the proof, note that $\boldsymbol{P}(z_0 \neq z'_0 \cap z_1 = z'_1) = \EE_{x \dist P_X} \EE_{x'\dist \hat{P}_{X|Z_n} \circ P_{Z_n|x}} \mathbb{1}(x \neq x')$ and proceed inductively.

\end{proof}

\subsection{Standard Rate-Distortion Theory}

Given a channel $P_{Z|X}$ rate distortion theory provides means of assessing lower bounds of the distortion of the channel by a function of the channels rate (maximum mutual information or capacity). For any prior $P_Y$, experiment $P_{X|Y}$, feature map $P_{Z|X}$, estimator $P_{A|Z}$ and loss function $L$ one defines the distortion 
$$
d = \EE_{y \dist P_{Y}} \EE_{x \dist P_{X|Y}} \EE_{z \dist P_{Z|X}} \EE_{z \dist P_{A|Z}} L(y,a)
$$
and rate
$$
R = \sup_{P_Z} I(X;Z).
$$
and rate-distortion function
$$
\phi_L(d)= \inf_{P_{A|Y}, \EE_{P_{YA}} L \leq d} I(Y;A) 
$$

i.e., the smallest mutual information of all channels $P_{A|Y}$ with distortion less than $d$.  The rate-distortion function is non-increasing, the higher the distortion the lower the required rate.

One obtains a lower bound of the distortion of the form $\phi_L^{-1}(R) \leq d$. $\phi$ is the rate-distortion function
$$
\phi_L(d)= \inf_{P_{A|Y}, \EE_{P_{YA}} L \leq d} I(Y;A) 
$$
Key to the rate distortion bound is that mutual information satisfies a data processing inequality, for a Markov chain
$$
Y \rightarrow X \rightarrow Z \rightarrow A
$$
$I(X;Z)\leq I(Y;A)$ \cite{Cover2012}. In particular this means for a Markov kernel of the form $P_{A|Y} = P_{A|Z} \circ P_{Z|X} \circ P_{X|Y}$ to have distortion less than $d$, 
$$
\phi_L(d) \leq I(X;Z) \leq R.
$$ 
This condition is necessary but not sufficient, leading to slack in the lower bound. Both the rate and the rate-distortion function can be computed via an iterative algorithm. We direct the reader to \cite{Cover2012} for derivations of the bound as well as the algorithm for calculating it. The major strength of this bound is that it applies \emph{for all} $P_Y$, $P_{X|Y}$,$P_{Z|X}$ and $P_{A|Z}$. If the marginal $P_X$ is known, the bound can be further tightened to 
$$
\phi_L^{-1}(I(X;Z)) \leq d.
$$ 
Rate distortion theory provides another justification of the use of mutual information as a surrogate for feature learning (different to theorem 4), and also provides means to assess how good a surrogate it is via the rate distortion function. On the following page are plots of the rate distortion curve for two different loss functions, firstly Brier loss and secondly the tilted Brier loss from the example in figure 2. From the plot one can see that more mutual information $I(X;Z)$ is required to have low distortion for the tilted Brier loss than the standard Brier loss. This is because the tilted Brier loss greatly penalizes mistakes made when classifying class 2, while penalizing other errors in a similar way to standard brier loss.
\begin{figure}[h]
\centering
\includegraphics[width=0.8\linewidth]{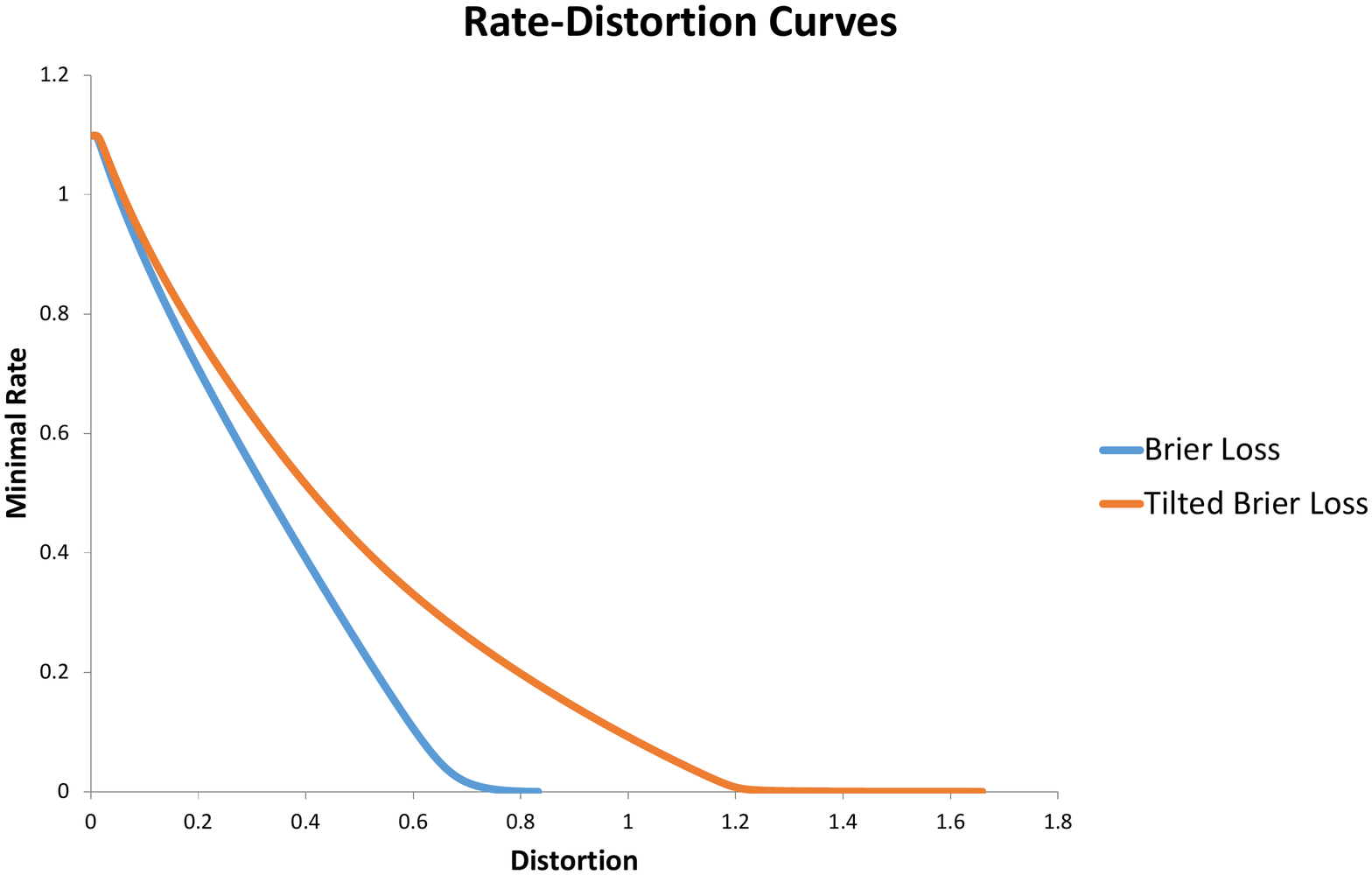}
\caption{Rate-Distortion Plots, see text}
\end{figure}

\subsection{Tighter Bounds via Generalized Rate Distortion Theory}

\begin{definition}
For convex $f: \RR^+ \rightarrow \RR$ with $f(1) = 0$, the \emph{$f$-information} of a joint distribution $P_{XY}$ is given by
$$
I_f(X;Y) = I_f(P_{XY}) = \EE_{P_{XY}} f(\frac{d(P_X \otimes P_Y)}{d P_{XY}}).
$$
\end{definition}
When $f(x) = -\log(x)$ we recover the mutual information. Much like mutual information, $f$-information also satisfies a data processing inequality. For any Markov chain
$$
Y \rightarrow X \rightarrow Z \rightarrow A
$$
$I_f(X;Z)\leq I_f(Y;A)$ \cite{Reid2009b}. As such one can use $f$-information to construct an alternative rate distortion function
$$
\phi_{L,f}(d)= \inf_{P_{A|Y}, \EE_{P_{YA}} L \leq d} I_f(Y;A) 
$$
and an alternative lower bound. Unlike the case of mutual information, there is not a fast iterative algorithm to calculate this function. However, it is easy to show that for fixed $d$ the above is a \emph{convex} optimization problem (as $f$-divergences are convex \cite{Reid2009b}).

\end{document}